# Pulmonary Embolism Mortality Prediction Using Multimodal Learning Based on Computed Tomography Angiography and Clinical Data

Zhusi Zhong, BS[a,b,c]*, Helen Zhang, BS[a,b]*, Fayez H. Fayad, BA[a,b]*, Andrew C. Lancaster, BS[d,e], John Sollee, BS[a,b], Shreyas Kulkarni, BS[a,b], Cheng Ting Lin, MD[d], Jie Li, PhD[c], Xinbo Gao, PhD[c], Scott Collins[a,b], Colin Greineder, MD[f], Sun H. Ahn, MD[a,b], Harrison X. Bai, MD[d], Zhicheng Jiao, PhD[a,b], Michael K. Atalay, MD, PhD[a,b]

*Co-first authors

[a] Department of Diagnostic Radiology, Rhode Island Hospital, Providence, RI, 02903, USA
[b] Warren Alpert Medical School of Brown University, Providence, RI, 02903, USA
[c] School of Electronic Engineering, Xidian University, Xi'an 710071, China
[d] Department of Radiology and Radiological Sciences, Johns Hopkins University School of Medicine, Baltimore, MD, 21205, USA
[e] Johns Hopkins University School of Medicine, Baltimore, MD, 21205, USA
[f] Department of Pharmacology, Medical School, University of Michigan, Ann Arbor, MI, 48109, USA

Zhusi Zhong: zhusi_zhong@brown.edu
Helen Zhang: helen_zhang2@brown.edu
Fayez H. Fayad: fayez_fayad@brown.edu
Andrew C. Lancaster: alancas4@jhmi.edu
John Sollee: john_sollee@brown.edu
Shreyas Kulkarni: shreyas_kulkarni@brown.edu
Cheng Ting Lin: clin97@jhmi.edu
Jie Li: leejie@mail.xidian.edu.cn
Xinbo Gao: xbgao@mail.xidian.edu.cn
Scott Collins: scollins1@lifespan.org
Colin Greineder: coling@med.umich.edu
Sun H. Ahn: sahn@lifespan.org
Harrison X. Bai: hbai7@jhu.edu
Zhicheng Jiao: zhicheng_jiao@brown.edu

**Abstract**

**Purpose:** Pulmonary embolism (PE) is a significant cause of mortality in the United States. The objective of this study is to implement deep learning (DL) models using Computed Tomography Pulmonary Angiography (CTPA), clinical data, and PE Severity Index (PESI) scores to predict PE mortality.

**Materials and Methods:** 918 patients (median age 64 years, range 13-99 years, 52% female) with 3,978 CTPAs were identified via retrospective review across three institutions. To predict survival, an AI model was used to extract disease-related imaging features from CTPAs. Imaging features and/or clinical variables were then incorporated into DL models to predict survival outcomes. Four models were developed as follows: (1) using CTPA imaging features only; (2) using clinical variables only; (3) multimodal, integrating both CTPA and clinical variables; and (4) multimodal fused with calculated PESI score. Performance and contribution from each modality were evaluated using concordance index (c-index) and Net Reclassification Improvement, respectively. Performance was compared to PESI predictions using the Wilcoxon signed-rank test. Kaplan-Meier analysis was performed to stratify patients into high- and low-risk groups. Additional factor-risk analysis was conducted to account for right ventricular (RV) dysfunction.

**Results:** For both data sets, the PESI-fused and multimodal models achieved higher c-indices than PESI alone.  Following stratification of patients into high- and low-risk groups by multimodal and PESI-fused models, mortality outcomes differed significantly (both $p<0.001$). A strong correlation was found between high-risk grouping and RV dysfunction.

**Conclusions:** Multiomic DL models incorporating CTPA features, clinical data, and PESI achieved higher c-indices than PESI alone for PE survival prediction.

**Abbreviations:**

AI = Artificial Intelligence

c-index = Concordance Index

CoxPH = Cox Proportional Hazards

CTPA = Computed Tomography Pulmonary Angiography

DL = Deep Learning

PE = Pulmonary Embolism

PESI = Pulmonary Embolism Severity Index

RSF = Random Survival Forest

RV = Right Ventricular

TTE, TEE = Transthoracic, Transesophageal Echocardiography

1. **Introduction**

Pulmonary embolism (PE) is a significant cause of morbidity and mortality, with nearly 600,000 cases and 60,000 deaths annually in the United States.[1,2] Following acute myocardial infarction and stroke, PE is the next most common cause of cardiovascular death in hospitalized patients.[3] Efficient diagnosis and management is key, as most deaths (>70%) occur within the first hour.[1] Clinical presentation is highly variable: common symptoms include tachycardia, dyspnea, and pleuritic chest pain.[4]

Given that timely and accurate risk stratification is vital for PE management, several prognostication tools have been developed. The Pulmonary Embolism Severity Index (PESI) is a well-validated tool that estimates 30-day mortality in patients with acute PE based on 11 clinical variables, serving as the present gold standard.[5] While PESI boasts a 99% negative predictive value of deterioration in patients classified as low-risk, positive predictive value for high-risk patients remains suboptimal at 11%.[6] Thus, there is a longstanding need to improve prognostication following diagnosis. Traditional survival methods include random survival forest (RSF), which employs a tree-based ensemble model, and Cox proportional hazards (CoxPH) models, which utilize hazard functions to estimate the linear impact of covariates on risk.[7,8]

With recent advances in artificial intelligence (AI), deep learning (DL)-based approaches have emerged as promising alternatives, significantly augmenting the interpretation of medical imaging studies.[9,10] AI-based models employed on computed tomography pulmonary angiography (CTPA) have been shown to diagnose PE with high accuracy and predict clot burden in acute cases.[11-14] Utilizing multimodal survival data, survival analysis techniques utilizing multimodal learning have exhibited enhanced robustness compared to single-modality techniques.[15,16] We hypothesized that a prognostication model combining imaging and clinical data would outperform PESI alone. We therefore aimed to develop and validate DL models using CTPA and clinical data to predict mortality in patients with PE.

## 2. Methods

This retrospective study was approved by the institutional review board (IRB) of INSTITUTION1, INSTITUTION2, and INSTITUTION3 with waiver of informed consent by each study participant.

The proposed clinical risk assessment algorithm is a DL-based framework incorporating multimodal neural networks embedded within an image analysis backbone model to predict survival outcomes (**Figure 1**).

We aimed to develop four deep learning-based prediction models capable of PE survival prediction for performance comparison:

- Model 1: Uses only CTPA imaging data (*deep imaging*)
- Model 2: Uses the 11 clinical variables considered within the PESI framework (*deep clinical*); notably, no manual weighting or scoring are applied to the data (no PESI score calculation)- the model is allowed to interpret these variables independently.
- Model 3: Incorporates both CTPA imaging data and aforementioned clinical variables (*deep multimodal*)
- Model 4: Combines the deep multimodal model with PESI score (*deep PESI-fused*).

*2.1. Clinical and Imaging Data Acquisition*

Retrospective chart review identified patients between March 2015 and February 2019 meeting the following inclusion criteria: confirmed PE on CTPA, transthoracic/transesophageal echocardiography (TTE, TEE) within two months of diagnosis. 918 patients were identified with corresponding clinical reports and CTPA series (3,978 CTPA acquisitions total). CTPA acquisitions for each patient were from the same date, consisting of different axial resolutions- the larger number of 3,978 represents the total sum of individual CTPA image acquisitions from all patients in our dataset. From electronic medical records, the 11 clinical variables considered in PESI were collected: age, sex, heart rate, systolic blood pressure, respiratory rate, temperature, mental status, previous pulmonary embolism or deep vein thrombosis, cancer, congestive heart failure, and chronic lung disease. Age was normalized within the overall dataset, and the remaining variables were binarized. Within the dataset, 94 patients (10.2%) had missing variables that required imputation using median values for binary variables and normalization for decimal variables. Mortality and hemodynamic collapse (as defined in PEITHO trial[17]) were recorded for applicable patients. PESI score was calculated for each patient using the aforementioned clinical variables.

Regarding ground truth for performance evaluation, clinical patient outcomes such as mortality and recorded time in electronic medical records were used to evaluate model performance with concordance index (c-index). For censored patients, the last recorded time point in the system was used as the cutoff time. C-index was employed to evaluate model performance by quantifying the concordance between predicted risk scores and observed survival times, taking censoring into consideration. This measures the likelihood of correctly ranking the survival times of pairs of individuals.

2.2. Image Preprocessing and PE Detection

A U-Net model was used for lung segmentation on CT to obtain lung region masks.[18] The corresponding lung regions of the respective CTPA images were extracted with a slice thickness of 1.25mm and scaled to 512x512 pixels. The entire image volume with $N$ slices was saved as a $N$x512x512 array. Hounsfield units of all slices were clipped to the range [−1000, 900] and zero-centered.

A robust trained PE detection model, PENet, was employed as the backbone for our image-based survival prediction model.[12] PENet automatically analyzes and identifies the most indicative features of PE within CTPAs. Across window-level predictions, the highest PE probability was used to determine a patient-level classification score. For each patient, the window-level prediction with the highest PE probability was selected, with the 2048 output features from the last convolutional layer designated as imaging features. In cases where a single patient had multiple CTPA acquisitions, the acquisitions were analyzed together to output a single set of optimal imaging features, resulting in each patient having only one corresponding set of imaging features.

*2.3. Learning-based Survival Analysis Framework*

Multimodal features included 11 PESI variables and extracted 2048-dimensional CTPA imaging features with highest PE probability formulated as $F^m, m \in [clin, img]$, used to train two independent survival prediction models. Survival prediction modules utilized a multilayer perceptron (MLP) with ReLU activation for feature encoding, followed by a linear regression layer with Sigmoid activation. The modules were trained with a Cox partial log-likelihood loss function on internal training and validation data, then used to build the survival prediction model.[9] Deep neural networks directly learned parameters from imaging and clinical covariates in a way that best modeled survival data, producing highly-individualized survival predictions.

Cross-modal fusion CoxPH models employed the two modal risk predictions to train a fused survival prediction model, incorporating time-to-event evaluation. This semi-parametric fusing enhances predictive capabilities by combining information from multiple modalities, offering a robust and nuanced understanding of survival outcomes. To assess performance of the multimodal learning-based model over PESI, a PESI-fused CoxPH model was evaluated, combining multimodal features and PESI. To compare relative performance between DL survival prediction and RSF, two single-modal RSFs were trained to model hazard functions, then fused with the same CoxPH models used in the DL survival framework.[7] Detailed methods and CLAIM checklist are included in **Supplemental Materials**.

*2.4. Statistical Analysis*

The INSTITUTION1 data were divided randomly into training, validation, and internal test sets (7:1:2). The INSTITUTION2/INSTITUTION3 data were consolidated as an external test set, due to relative similarity and smaller size. The survival prediction models underwent training and validation using the INSTITUTION1 training set and validation dataset. The frameworks were then applied to the internal and external test set. Model performance was evaluated by c-index and compared to PESI predictions using the Wilcoxon signed-rank test.[19] Net Reclassification Improvement (NRI) was used to assess performance improvement from each modality in mortality classification prediction.[20] Kaplan-Meier analysis was performed to stratify patients into high- and low-risk groups.[21] CoxPH models were used to analyze risk scores from the multimodal models.[8] Significance level was set to $p<0.05$.

*2.5. RV Dysfunction Factor-risk Analysis*

As right ventricular (RV) dysfunction is an important prognostic factor in PE, we took it into consideration when evaluating the survival framework.[4,22,23] First, a binarized label for RV dysfunction was collected for each patient. This was then incorporated into a factor-risk analysis with multimodal survival predictions. Patients were sorted by the median of multimodal predicted risk. RV dysfunction was then designated as a risk factor.

**3. Results**

*3.1. Subjects and Clinical Outcomes*

A total of 918 PE patients (median age 64 years, 52% female) and 3,978 CTPAs were identified, with an average of 4 same-day CTPAs per patient. The INSTITUTION1 dataset included 485 patients, while the pooled INSTITUTION2-INSTITUTION3 dataset included 433 patients. 163 patients were deceased at time of review. Sixty-five patients expired within 30 days of diagnosis, and 31 expired within seven days of diagnosis. Furthermore, 77 patients suffered hemodynamic collapse within seven days of diagnosis. Detailed clinical summary is shown in **Table 1**.

*3.2. PE Detection Performance*

The PENet backbone achieved an accuracy of 0.985 and AUROC of 0.971 in identifying PE on the overall dataset. Furthermore, it achieved a sensitivity of 0.941, a specificity of 1.000, a precision of 1.000, and a F1 score of 0.985. Class Activation Maps (CAMs) were utilized to visualize the neural network's regions of interest, extracted from the final convolutional layer of PENet and weighted by learned PE classification layer. Highlighted regions of the window prediction indicate predicted locations of PE (**Figure 2**). Ultimately, the PENet image-based analysis network effectively captured PE-related features in CTPA.

*3.3. Overall Survival Prediction Performance*

DL survival analysis frameworks were based on (a) CTPA imaging data, (b) clinical variables, (c) multimodal prediction incorporating both CTPA data and clinical variables, and (d) multimodal model fused with PESI score, as previously established. Model performance is visualized in **Figure 3**. As a comparative baseline, PESI was evaluated alone. For both internal and external data sets, the PESI-fused model achieved higher c-indices than PESI alone (**Table 2**). Following stratification of patients into high- and low-risk groups by the PESI-fused model, Kaplan-Meier analysis revealed significantly different mortality outcomes (p<0.001), shown in **Figure 4**.[24]

*3.4. Short-term Survival Prediction Performance*

As PESI estimates risk of 30-day mortality, short-term survival was compared by truncating time-to-event labels at a 30-day maximum.[5] PESI demonstrated greater performance in predicting short-term PE survival compared to long-term (**Table 3**). However, multimodal and PESI-fused models still exhibited significant performance improvement over PESI in short-term survival prediction.

*3.5. Feature Importance*

For the clinical survival prediction model, a summary of the predictive ability of each clinical feature and respective feature importance within the model is illustrated in **Figure 5**. Predictive ability measured each variable's contribution to model performance, while feature importance was determined through coefficients of feature selection with the learning-able neuron weights. Age and history of cancer had the greatest predictive ability, while history of cancer had the greatest feature importance.

*3.6. Multimodal Improvement*

The multimodal learning framework integrated survival characteristics from multiple modalities- to analyze the individual contributions of each modality, as well as the contribution of PESI to the PESI-fused model, we used NRI to evaluate the accuracy improvement achieved by incorporating each.[20] Predicted risk probabilities were binarized with a threshold of 0.7 to obtain predicted mortality categories. We evaluated NRI by calculating risk scores between (a) imaging and multimodal, (b) clinical and multimodal, and (c) multimodal and PESI-fused models, represented as +Clinical, +Imaging, and +PESI, respectively (**Table 4**). The positive values of +Clinical and +Imaging indicate that both clinical and imaging data contributed to improved predictive performance of the multimodal framework.

*3.7. RV Dysfunction and PE Risk Classification*

We identified RV dysfunction as a risk factor in 16 out of 433 patients in the external test set. We visualized the positioning of the 16 patients in **Figure 6a**, with the multimodal survival framework identifying 68.8% of RV dysfunction patients as high-risk. The multimodal survival prediction model also demonstrated a high correlation between high-risk identification and mortality, as shown in **Figure 6b**. Fifty-five of the 65 mortality patients were predicted as high-risk, yielding a mortality classification accuracy of 84.6%.

**4. Discussion**

*4.1. Rationale for Study Approach*

Prior studies have explored the potential of AI-based models in PE assessment, focusing on improving detection and diagnosis of PE.[11,25-27] Once a diagnosis of acute PE has been made, determining disease severity is important to guiding clinical management. PESI is a well-validated risk assessment tool for prediction of 30-day morbidity and mortality, commonly used in clinical practice.[5] In a meta-analysis including 71 studies and 44,298 patients, PESI and simplified PESI tools were the most highly-validated models available.[28] However, PESI's positive predictive value for high-risk patients is only 11%.[6] Considering PESI's limitations, we sought to develop AI-based models to build upon existing tools and improve prognostication.

While other studies have shown great potential for AI in detecting and diagnosing PE, few have shown benefits of AI for prognostication. Additionally, incorporation of multimodal data allows for more heterogeneous analysis. Somani et al. supported that use of fusion models may outperform non-fusion models in PE detection, supporting our efforts to assess efficacy of different permutations of fusion models in prognostication of PE.[25] A recent study explored the use of a multimodal model for PE risk stratification, based on prediction of thirty-day all-cause mortality.[29] A deep neural network (TabNet) was combined with a CNN, relying on a single binary label for each CTPA scan. Their fusion model achieved higher performance (AUC: 0.96) compared to clinical (0.87) and imaging (0.82) models. Our study further supports how multimodal approaches can improve healthcare decision-making and prognostication in PE patients.

*4.2. Discussion of Study Findings*

In this study, we showed that DL models incorporating combined imaging and clinical features can achieve high performance in predicting PE mortality, improving performance over PESI alone. The multimodal model outperformed both imaging and clinical models, indicating enhanced robustness from combining imaging and clinical data. The PESI-fused model slightly outperformed the multimodal model, indicating marginal benefit from incorporating the PESI framework. Models were also compared to RSF, with RSF outperforming the imaging model, clinical model, and PESI on the internal test set. However, RSF outperformed only the imaging model on the external test set. On both internal and external test sets, RSF was outperformed by the deep multimodal and PESI-fused models, demonstrating benefits of deep multimodal learning over a traditional survival method.

Given that PESI estimates the risk of 30-day mortality, additional survival comparison was conducted to evaluate 30-day performance. PESI demonstrated greater performance in predicting short-term PE survival compared to long-term, consistent with its clinical purpose. The clinical, multimodal, and PESI-fused models demonstrated improved performance in short-term prediction compared to long-term on the internal test set. However, they demonstrated lower performance compared to long-term on the external test set. Despite the improved performance of PESI in short-term prediction, the majority of DL models still demonstrated higher performance. On internal testing, clinical, multimodal, and PESI-fused models achieved higher c-indices than PESI. On external testing, PESI outperformed the clinical model but underperformed the multimodal and PESI-fused models. These findings indicate PESI's performance is improved in short-term prediction. However, the deep multimodal and PESI-fused models still demonstrate improved performance, subject to model generalizability. This provides insight into how model performance may vary based on the specific outcome being assessed, as short-term mortality may be influenced less by competing risk factors.

For the clinical survival prediction model, we identified the predictive ability and importance of each feature. Age and history of cancer were found to have the greatest predictive ability. History of cancer had the greatest feature importance. This analysis may provide valuable insight into the underlying mechanisms or risk factors related to the predicted outcome. The alignment of our survival prediction model with observations in clinical practice provides further validation of model rationality.

NRI was used to analyze the contributions of different modalities to the multimodal framework, as well as the contribution of PESI to the PESI-fused model, by measuring the accuracy improvement achieved by incorporating each. The NRI values for +Clinical and +Imaging were positive, indicating improved performance from the incorporation of clinical and imaging data in the multimodal framework. Meanwhile, the values for +PESI were negative/lower, indicating less of a contribution. This suggests that the integration of imaging and clinical variables provides valuable and complementary information for survival prediction, resulting in more refined and reliable classification of individuals. Much of the information within PESI is already included in clinical variables, but conflicting characterization performance of PESI may lessen its contribution to the PESI-fused model.

Given the importance of RV dysfunction as a risk factor in PE patients, an additional factor-risk analysis was performed with the multimodal survival predictions. The multimodal survival model identified 68.8% of RV dysfunction patients as high-risk. The model also demonstrated a high correlation between high-risk identification and mortality, identifying 84.6% of mortality patients as high-risk. Through this risk stratification, the survival model was shown to be capable of predicting mortality, as well as having a relatively strong correlation with the prognostic factor of RV dysfunction. Thus, our model validates the association between RV dysfunction and death in PE patients.

*4.3. Limitations*

There are several limitations to this study. Like most DL-based survival analysis models, there is a concern for generalizability given that the model was trained using limited data from a single institution. To ensure external validity and generalizability of our models, we trained and validated them first on the single-institution internal dataset, then tested their accuracy on the previously-unseen multi-institution external dataset. Additionally, concatenation was used to fuse the two survival prediction branches- a more effective feature fusion mechanism between imaging and clinical data remains to be investigated. As we did not have access to data regarding patient treatment strategies within hospitals, we were not able to take clustering in treatment approaches into account. We were not able to compare the predictive value of the models between different settings (inpatient, emergency department, outpatient). Lastly, due to the CTPA requirement within our inclusion criteria, our study excludes more severe cases and perhaps the majority of PE mortality as these patients typically do not survive long enough to undergo CTPA.

Before being widely accepted, our models will likely require additional validation on larger and more diverse datasets, as well as prospective testing of the developed models. With the application of DL into medical care, appropriate and robust regulatory measures must be passed, and radiologists/clinicians will need to be trained to implement such models into their workflows.

## 5. Conclusions

Multiomic DL models based on combined CTPA features and clinical variables demonstrated improved performance compared to PESI score alone for mortality prediction in PE. The addition of PESI to the multimodal model demonstrated only a marginal performance improvement, illustrating that AI-based models are sufficiently capable of survival prediction. The multimodal models similarly improved performance upon PESI alone in 30-day mortality risk estimation. Through NRI analysis, clinical and imaging data were both independently shown to contribute to improved performance of the multimodal model. These findings demonstrate the strength of a multimodal DL model in comparison to the current clinical standard of PESI, turning prognosis into an intelligent process that integrates greater clinical and imaging information. Additionally, we demonstrated concordance of our model with clinical indicators of mortality, such as RV dysfunction. Further analysis can shed more light on the connectedness of various risk factors with mortality in PE patients, and how this information can be leveraged for model development in survival prediction. However, the benefits of our model can only be confirmed by additional validation on larger and more diverse datasets, as well as prospective testing of the developed models.

Our study highlights the utility of DL-based models in prognostication and risk stratification in patients with PE. AI has the potential to improve the clinical workflow for radiologists and clinicians by providing rapid and accurate diagnostic and prognostic information. By offering timely yet accurate risk stratification for PE patients, AI may offer a substantial benefit to patients and providers by informing clinical decision-making, potentially improving patient outcomes.

**Acknowledgments**

None.

**Figure Legends:**

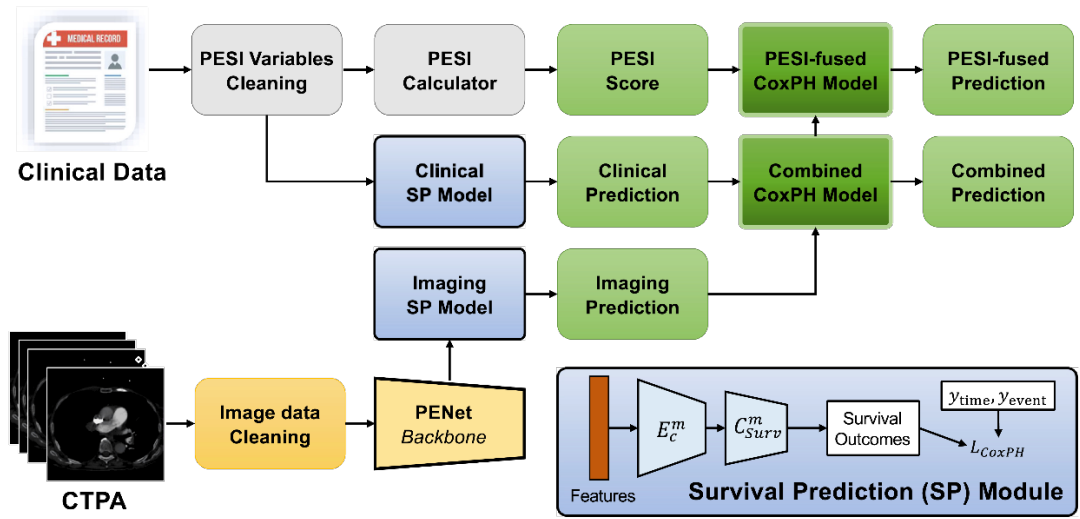

**Figure 1. Data Analysis Workflow.** This Central Illustration provides an overview of the data analysis workflow, including the proposed Pulmonary Embolism (PE) deep survival analysis framework.

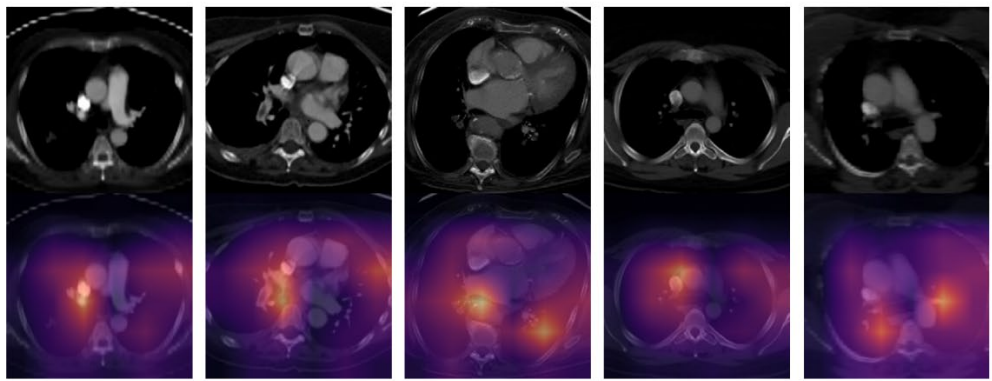

**Figure 2. Class Activation Maps (CAMs).** Class activation maps (CAMs) highlight the image areas most important for PE detection model decision-making.

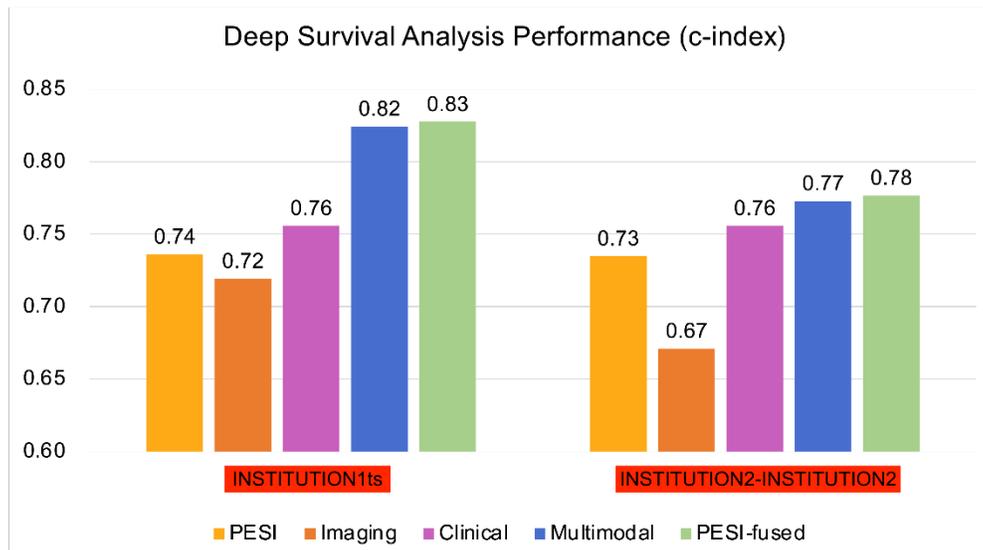

**Figure 3. Performance of deep survival analysis models.** Comparison of deep survival analysis models' overall performance on different testing datasets.

PESI = Pulmonary Embolism Severity Index. INSTITUTION1ts = internal test set. INSTITUTION2-INSTITUTION3 = external test set.

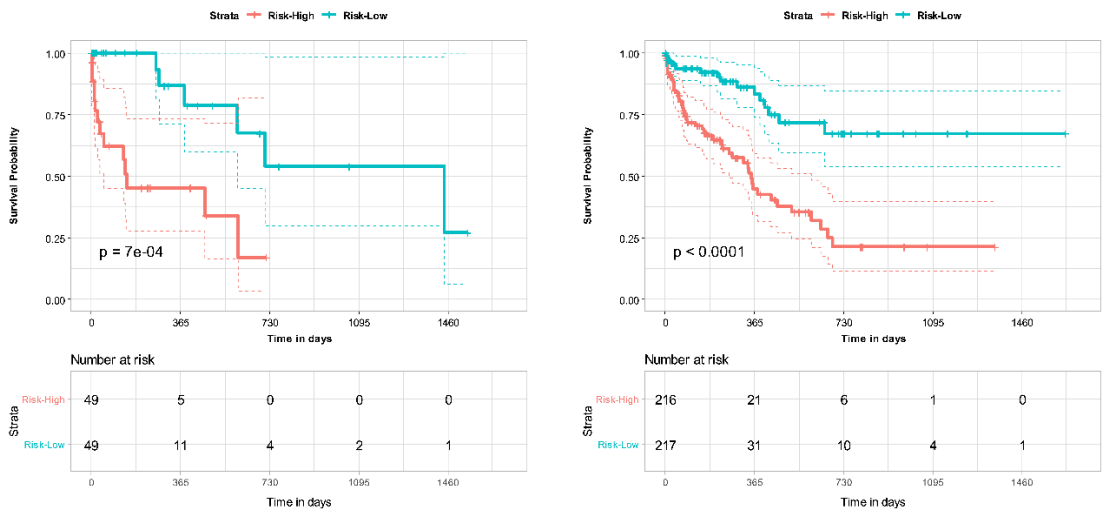

**Figure 4**. **Kaplan-Meier curves.** Kaplan-Meier curves for INSTITUTION1ts (left) and INSTITUTION2- INSTITUTION3 (right) with patients stratified into high- and low-risk groups by the PESI-fused model.

INSTITUTION1ts = internal test set. INSTITUTION2-INSTITUTION3 = external test set.

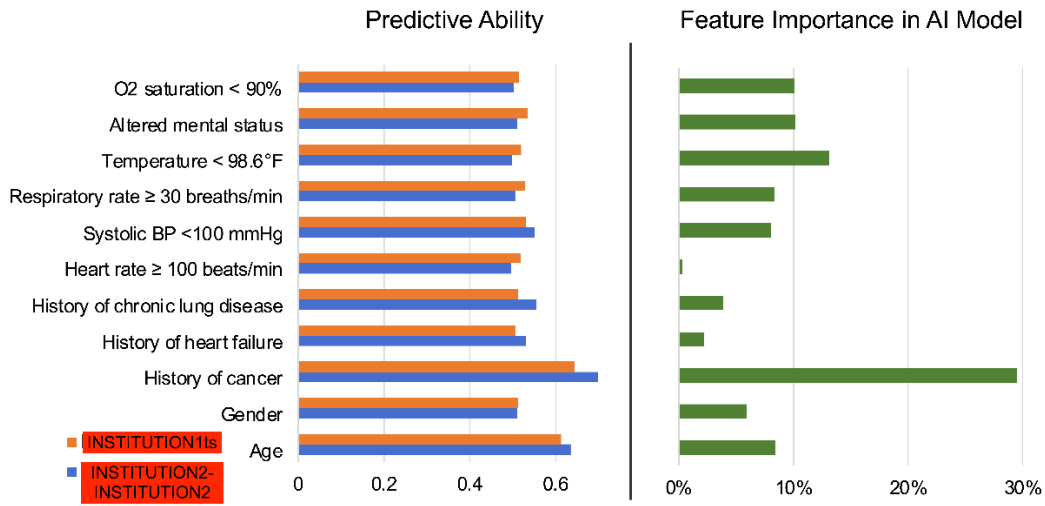

**Figure 5. Feature Importance.** Predictive ability of each clinical feature (left) and feature importance in AI model (right).

INSTITUTION1ts = internal test set. INSTITUTION2-INSTITUTION3 = external test set.

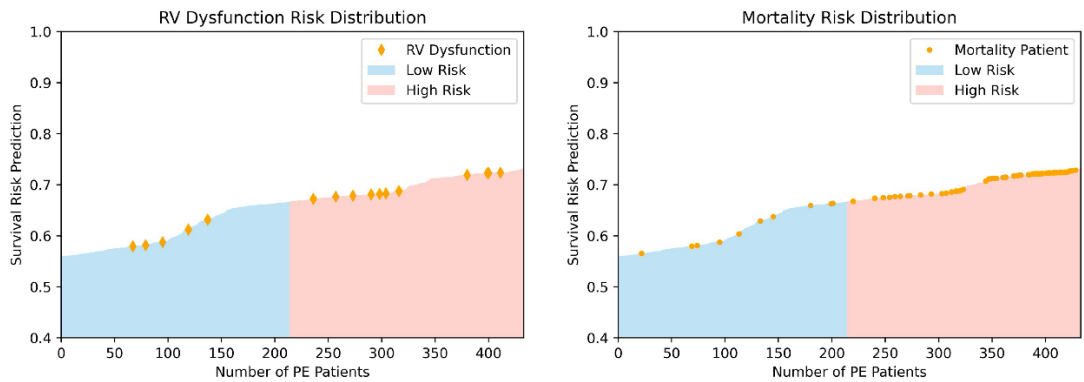

**Figure 6. Predicted risk distribution of external testing set.** Figure (a) showcases 16 patients with RV dysfunction, 68.8% of which are high-risk, and Figure (b) demonstrates a high correlation between high-risk identification and mortality. (a) Diamonds represent PE patients with RV dysfunction. (b) Triangles represent mortality.

**Table 1. Patient characteristics.**

|  | **Internal set (n=485)** | **External set (n=433)** | **P value** |
|---|---|---|---|
| Age (years) | 62 (24) | 67 (28) | **0.004** |
| Male | 234 (48%) | 201 (46%) | 0.560 |
| Death | 98 (20%) | 65 (15%) | **0.038** |
| Chronic cancer | 159 (32%) | 103 (24%) | **0.002** |
| Chronic heart failure | 32 (7%) | 38 (8%) | 0.218 |
| Chronic obstructive pulmonary disease | 105 (22%) | 90 (21%) | 0.737 |
| Heart rate ≥ 110 beats/min | 86 (18%) | 63 (15%) | 0.186 |
| Systolic BP < 100 mmHg | 46 (9%) | 40 (9%) | 0.890 |
| Respiratory rate ≥ 30 breaths/min | 11 (2%) | 13 (3%) | 0.490 |
| Temperature < 98.6°F | 25 (5%) | 16 (4%) | 0.283 |
| Altered mental status | 48 (10%) | 18 (4%) | **0.001** |
| O2 saturation < 90% | 14 (3%) | 10 (2%) | 0.581 |
| PESI | 88 (46) | 85 (45) | 0.115 |

Detailed patient characteristics of PESI clinical variables used to calculate PESI score for each patient.

All continuous variables are reported as median (interquartile range), and all categorical variables are reported as number (%). Statistically significant p-values are bolded (p < 0.05). Deceased status is not a PESI clinical variable.

BP = Blood Pressure. PESI = Pulmonary Embolism Severity Index.

**Table 2. Overall survival prediction performance.**

| Dataset | PESI | RSF Multimodal | Deep Imaging | Deep Clinical | Deep Multimodal | Deep PESI-fused |
|---|---|---|---|---|---|---|
| INSTITUTION1 tr | 0.706 (0.702 to 0.709) | 0.978 (0.977 to 0.978) | 0.786 (0.775 to 0.796) | 0.768 (0.765 to 0.773) | 0.835 (0.831 to 0.840) | 0.838 (0.833 to 0.843) |
| INSTITUTION1 ts | 0.736 (0.724 to 0.746) | 0.787 (0.777 to 0.798) | 0.719 (0.712 to 0.730) | 0.756 (0.743 to 0.768) | 0.824 (0.816 to 0.834) | 0.827 (0.819 to 0.837) |
| INSTITUTION2 | 0.734 (0.731 to 0.737) | 0.698 (0.693 to 0.700) | 0.674 (0.671 to 0.677) | 0.765 (0.762 to 0.768) | 0.782 (0.779 to 0.785) | 0.786 (0.783 to 0.790) |
| INSTITUTION3 | 0.686 (0.661 to 0.707) | 0.536 (0.504 to 0.561) | 0.625 (0.668 to 0.673) | 0.609 (0.585 to 0.634) | 0.658 (0.635 to 0.678) | 0.664 (0.640 to 0.685 |
| INSTITUTION2 & INSTITUTION3 | 0.735 (0.731 to 0.737) | 0.678 (0.674 to 0.681) | 0.671 (0.668 to 0.674) | 0.756 (0.752 to 0.758) | 0.773 (0.770 to 0.775) | 0.777 (0.774 to 0.779) |

Overall c-index values and corresponding 95% confidence intervals of PESI and prediction models.

INSTITUTION3 = INSTITUTION3. PESI = Pulmonary Embolism Severity Index. RSF = Random Survival Forest. INSTITUTION1 = INSTITUTION1. INSTITUTION1tr = training set. INSTITUTION1ts = internal test set. INSTITUTION2 = INSTITUTION2. INSTITUTION2-INSTITUTION3 = external test set.

**Table 3**. **Short term survival prediction performance.**

| Dataset | PESI | Deep Imaging | Deep Clinical | Deep Multimodal | Deep PESI-fused |
|---|---|---|---|---|---|
| INSTITUTION1tr | 0.724 (0.722 to 0.727) | 0.830 (0.829 to 0.831) | 0.696 (0.694 to 0.699) | 0.778 (0.777 to 0.780) | 0.781 (0.779 to 0.783) |
| INSTITUTION1ts | 0.757 (0.755 to 0.760) | 0.705 (0.703 to 0.708) | 0.781 (0.777 to 0.784) | 0.828 (0.826 to 0.831) | 0.837 (0.835 to 0.839) |
| INSTITUTION2 | 0.738 (0.736 to 0.741) | 0.609 (0.606 to 0.612) | 0.749 (0.746 to 0.751) | 0.757 (0.754 to 0.760) | 0.762 (0.759 to 0.764) |
| INSTITUTION3 | 0.857 (0.855 to 0.858) | 0.573 (0.571 to 0.576) | 0.379 (0.376 to 0.382) | 0.742 (0.739 to 0.744) | 0.757 (0.755 to 0.760) |
| INSTITUTION2 & INSTITUTION3 | 0.754 (0.751 to 0.756) | 0.608 (0.606 to 0.611) | 0.738 (0.735 to 0.741) | 0.761 (0.758 to 0.763) | 0.765 (0.762 to 0.767) |

Short term (30-day) survival prediction performance as measured by c-index values and corresponding 95% confidence intervals of PESI and prediction models.

INSTITUTION3 = INSTITUTION3. PESI = Pulmonary Embolism Severity Index. INSTITUTION1 = INSTITUTION1. INSTITUTION1tr = training set. INSTITUTION1ts = internal test set. INSTITUTION2 = INSTITUTION2. INSTITUTION2- INSTITUTION3 = external test set.

**Table 4**. **Net Reclassification Improvement (NRI) analysis.**

| Dataset | +Clinical | +Imaging | +PESI |
|---|---|---|---|
| INSTITUTION1tr | 0.168 (0.079 to 0.258) | 0.156 (0.092 to 0.223) | 0.000 (-0.026 to 0.017) |
| INSTITUTION1ts | 0.137 (-0.024 to 0.299) | 0.103 (0.037 to 0.191) | -0.026 (-0.118 to 0.025) |
| INSTITUTION2 | 0.142 (0.065 to 0.232) | 0.064 (0.019 to 0.110) | 0.000 (-0.023 to 0.016) |
| INSTITUTION3 | -0.162 (-0.347 to -0.000) | -0.092 (-0.313 to 0.013) | 0.006 (0.000 to 0.020) |
| INSTITUTION2 & INSTITUTION3 | 0.126 (0.037 to 0.207) | 0.058 (0.011 to 0.101) | 0.003 (-0.018 to 0.017) |

Risk scores were calculated between imaging and multimodal (+Clinical), clinical and multimodal (+Imaging), and multimodal and PESI-fused (+PESI) models for each dataset.

INSTITUTION3 = INSTITUTION3. PESI = Pulmonary Embolism Severity Index. INSTITUTION1 = INSTITUTION1. INSTITUTION1tr = training set. INSTITUTION1ts = internal test set. INSTITUTION2 = INSTITUTION2. INSTITUTION2- INSTITUTION3 = external test set.